\documentclass[11pt,a4paper]{article}
\usepackage[hyperref]{acl2021}
\usepackage{times}
\usepackage{latexsym}
\usepackage{graphicx}
\usepackage{booktabs}
\usepackage{url}
\usepackage{multirow}
\usepackage{amsmath}
\usepackage{amssymb}

\usepackage[T1]{fontenc}
\usepackage{pbox}
\usepackage{xcolor}
\usepackage[colorinlistoftodos,prependcaption,textsize=footnotesize]{todonotes}

\usepackage[nameinlink]{cleveref}

\crefformat{section}{\S#2#1#3} 
\crefformat{subsection}{\S#2#1#3}
\crefformat{subsubsection}{\S#2#1#3}

\usepackage{microtype}

\aclfinalcopy 

\newenvironment{itemize*}%
  {\begin{itemize}%
    \setlength{\itemsep}{0.9pt}%
    \setlength{\parskip}{0.9pt}%
    \setlength{\topsep}{0.9pt}}%
  {\end{itemize}}

\newcommand{\symfootnote}[1]{%
\let\oldthefootnote=\thefootnote%
\stepcounter{mpfootnote}%
\addtocounter{footnote}{-1}%
\renewcommand{\thefootnote}{\fnsymbol{mpfootnote}}%
\footnote{#1}%
\let\thefootnote=\oldthefootnote%
}

\title{BERT Busters: Outlier Dimensions that Disrupt Transformers}
\newcommand*{\affaddr}[1]{#1} 
\newcommand*{\affmark}[1][*]{\textsuperscript{#1}}
\newcommand*{\email}[1]{\texttt{#1}}

\date{}

\begin{document}
\author{
Olga Kovaleva\thanks{\hspace{0.2cm}Authors contributed equally to this work.} \affmark[1], Saurabh Kulshreshtha\footnotemark[1]  \affmark[1], Anna Rogers\affmark[2] and Anna Rumshisky\affmark[1]\\
\affaddr{\affmark[1]Department of Computer Science, University of Massachusetts Lowell}\\
\affaddr{\affmark[2]Center for Social Data Science, University of Copenhagen} \\
\affmark[1]\email{\{okovalev,skul,arum\}@\{cs.uml.edu\}}\\
\affmark[2]\email{arogers@sodas.ku.dk}\\

}
\maketitle
\begin{abstract}
Multiple studies have shown that Transformers are remarkably robust to pruning. Contrary to this received wisdom, we demonstrate that pre-trained Transformer encoders are surprisingly fragile to the removal of a very small number of features in the layer outputs ($<$0.0001\% of model weights). In case of BERT and other pre-trained encoder Transformers, the affected component is the scaling factors and biases in the LayerNorm. The outliers are high-magnitude normalization parameters that emerge early in pre-training and show up consistently in the same dimensional position throughout the model. We show that disabling them significantly degrades both the MLM loss and the downstream task performance. This effect is observed across several BERT-family models and other popular pre-trained Transformer architectures, including BART, XLNet and ELECTRA; we also show a similar effect in GPT-2.
\end{abstract}

\section{Introduction}

Pre-trained Transformer-based models  \cite{vaswani2017attention} have become widely popular in a variety of NLP applications. 
%
%
Multiple studies of BERT-family models \cite{devlin2019bert} 
showed that Transformers are remarkably robust to pruning \cite{gordon2020compressing,PrasannaRogersEtAl_2020_When_BERT_Plays_Lottery_All_Tickets_Are_Winning,ChenFrankleEtAl_2020_Lottery_Ticket_Hypothesis_for_Pre-trained_BERT_Networks,michel2019sixteen}. This work presents a different and unexpected result: it is possible to dramatically disrupt the performance of BERT and other Transformer-based architectures by modifying very few weights (less than 0.0001\% for BERT).

In particular, we show that there is a very small number of outlier dimensions that regularly appear in the same position 
in the pre-trained encoder layers of a given Transformer model.
We demonstrate that this effect holds for different Transformer-family architectures, including multiple variants of BERT, as well as ELECTRA \cite{clark2020electra}, BART \cite{lewis2019bart}, and XLNet \cite{yang2019xlnet}. A similar phenomenon is also present in the decoder layers of GPT-2 \cite{radford2019language}.  
When these dimensions are disabled throughout the model in the concluding transformation of each layer, they can drastically reduce the overall model performance. 
With the exception of GPT-2, the last transformation in each layer of these models is normalization (LayerNorm), which is what we mainly focus on in this study.

The contributions of this work are as follows:
\begin{itemize*}
\item We identify certain outlier dimensions in Transformer layer outputs and show that they play a crucial role in both language modeling and downstream task performance. Disabling the weights for these output dimensions drastically degrades performance (up to 44 points).

\item We show that this effect holds for the encoder layers of six different models of the BERT family, as well as other popular pre-trained Transformer-based models including ELECTRA, BART, and XLNet. In GPT-2, a similar phenomenon is observed in the output dense transformation of the decoder layers.

\item{We demonstrate that outlier weights emerge gradually and begin to emerge early in pre-training, causing abnormal spikes at select dimensions in the output embedding vectors.}

\end{itemize*}

To our knowledge, this is the first work to establish the presence of very few regular outliers in the output Transformer representations and their importance for the model performance. It is not clear why these features emerge, but the final transformations clearly play a larger role in the Transformer layers than is usually assumed, and this needs further investigation.

This paper is organized as follows. After a brief overview of related work (\cref{sec:related}), we introduce the methodology for defining, locating, and disabling the BERT outlier weights in \cref{sec:methodology}. In  \cref{sec:mlm} and \cref{sec:glue}, we quantify the effect of disabling these weights
both in pre-training and in downstream tasks. 
In \cref{sec:others}, we demonstrate that other Transformers (BART, ELECTRA, XLNet, and GPT-2) also exhibit similar behavior. \cref{sec:ablation} evaluates magnitude- and position-based criteria for identifying the outlier dimensions and compares them with our proposed criteria.
In \cref{sec:pretraining}, we replicate the outlier effect in a BERT model during pretraining and study its dynamics.

\section{Related work}
\label{sec:related}

\paragraph{Transformer layer outputs.}
At a high level, Transformer encoder layers consist of multi-head self-attention followed by a dense layer \citet{vaswani2017attention}. Most contemporary Transformers use 
normalization to improve the speed and stability of training. 
%

Usually, the outputs of both self-attention and linear layers undergo the layer normalization transformation (LayerNorm, \citet{ba2016layer}). Each LayerNorm transformation is parameterized by a separate set of learned weights (\textit{scaling factors} and \textit{biases}). \citet{xiong2020layer} refer to this configuration as \textit{post-LN}. In the \textit{pre-LN} variant adopted by the GPT-2 model, LayerNorm is applied prior to the self-attention or linear transformations instead.

We will refer to the outputs of the final transformation in the encoder layer as \textit{features} and the parameters of this transformation as \textit{weights}. The final transformation is LayerNorm for 
all models considered in this study except GPT-2, where the last component is a MLP.


\begin{figure}[t]
    \centering
    \includegraphics[width=0.5\linewidth]{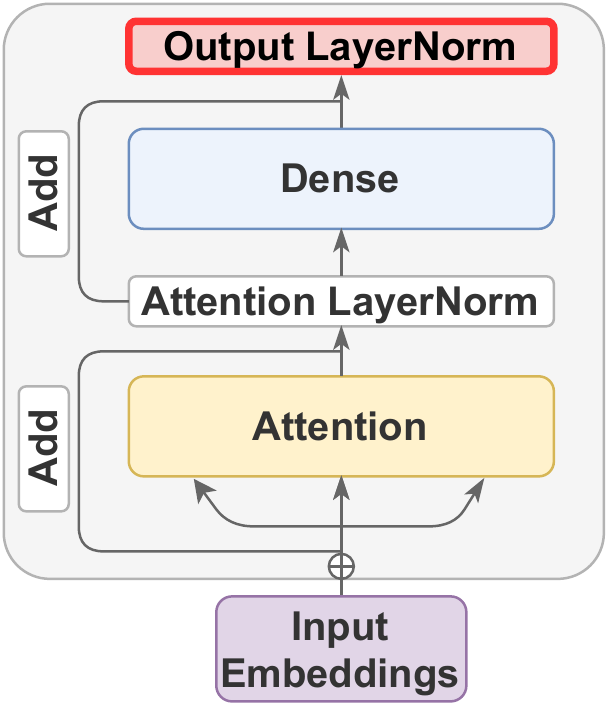}
    \caption{Transformer encoder layer, adapted from \cite{vaswani2017attention}.}
    \label{fig:bert}
\end{figure}

Like other normalization techniques, LayerNorm operates in two steps.
For a given input $x_{i}$ of the $i$-th layer with a hidden dimension $m$, LayerNorm computes 
mean and variance across the features:
\begin{equation}
    \mu_i = \frac{1}{m} \sum_{j=1}^{m} x_{ij}, \; \sigma_i^2 = \frac{1}{m} \sum_{j=1}^m \left(x_{ij} - \mu_i\right)^2
    \label{eq:mu_sigma}
\end{equation}
The inputs are then normalized and a learnable scale-shift transformation is applied to produce the normalized output embedding:
\begin{equation}
    \hat{x}_{ij} = \frac{x_{ij} - \mu_i}{\sqrt{\sigma_i^2 + \epsilon}}, \; y_i = \gamma \odot \hat{x}_i + \beta
    \label{eq:scale_shift}
\end{equation}
where $\gamma\in\mathbb{R}^m$ and $\beta\in\mathbb{R}^m$ are trainable parameters referred to as \textit{scaling factor} and \textit{bias} (shift). 

So far there have been few studies of normalization strategies in Transformer architectures and they focused mostly on the description of the training process. \citet{xiong2020layer} 
show that the location of LayerNorm in Transformer affects the gradient flow and demonstrate the need of the warmup stage.
\citet{nguyen2019transformers} 
inject multiple normalization blocks in specific network submodules to improve model performance. More recently, LayerNorm alternatives have been proposed and shown to have better gradient propagation through the network \cite{NEURIPS2019_2f4fe03d,shen2020powernorm}. 

\paragraph{Overparametrization.}

After the initial reports of redundancy in the BERT model 
\cite{michel2019sixteen,KovalevaRomanovEtAl_2019_Revealing_Dark_Secrets_of_BERT}, compressing Transformers quickly became a subfield of its own  \cite{model:tinyBERT,model:8bitbert,FanGraveEtAl_2019_Reducing_Transformer_Depth_on_Demand_with_Structured_Dropout,GuoRushEtAl_2020_Parameter-Efficient_Transfer_Learning_with_Diff_Pruning}. See overviews by \citet{ganesh2020compressing} and \citet{RogersKovalevaEtAl_2020_Primer_in_BERTology_What_We_Know_About_How_BERT_Works}.

Pruning is a class of methods for model compression which involves setting some of its weights to zeros with minimal performance loss. Much pruning work focuses on compression for the sake of efficiency, but it is also used for model analysis, and that is our goal as well. The most common approach is selecting the weights to be pruned by magnitude \cite{NIPS2015_ae0eb3ee}. 

Some of the recent findings are that the lottery ticket hypothesis \cite{FrankleCarbin_2019_Lottery_Ticket_Hypothesis_Finding_Sparse_Trainable_Neural_Networks} holds for BERT: its largest weights do form subnetworks that can be retrained alone to reach the performance close to that of the full model \cite{PrasannaRogersEtAl_2020_When_BERT_Plays_Lottery_All_Tickets_Are_Winning,ChenFrankleEtAl_2020_Lottery_Ticket_Hypothesis_for_Pre-trained_BERT_Networks,gordon2020compressing}. In structured pruning, the best subnets of BERT's heads and MLPs (selected by importance scores) do not quite reach the full model performance, but the worst ones are still much better than the worst magnitude-based subnets \cite{PrasannaRogersEtAl_2020_When_BERT_Plays_Lottery_All_Tickets_Are_Winning}, presumably because they retain a lot of high-magnitude weights. 

\section{Outlier weights in BERT models} \label{sec:methodology}
In this section, we introduce the methodology for identifying outlier dimensions and describe our method for disabling these outlier features.

\subsection{Identification}
\label{definition}

\begin{figure} 
    \centering
    \includegraphics[width=\linewidth]{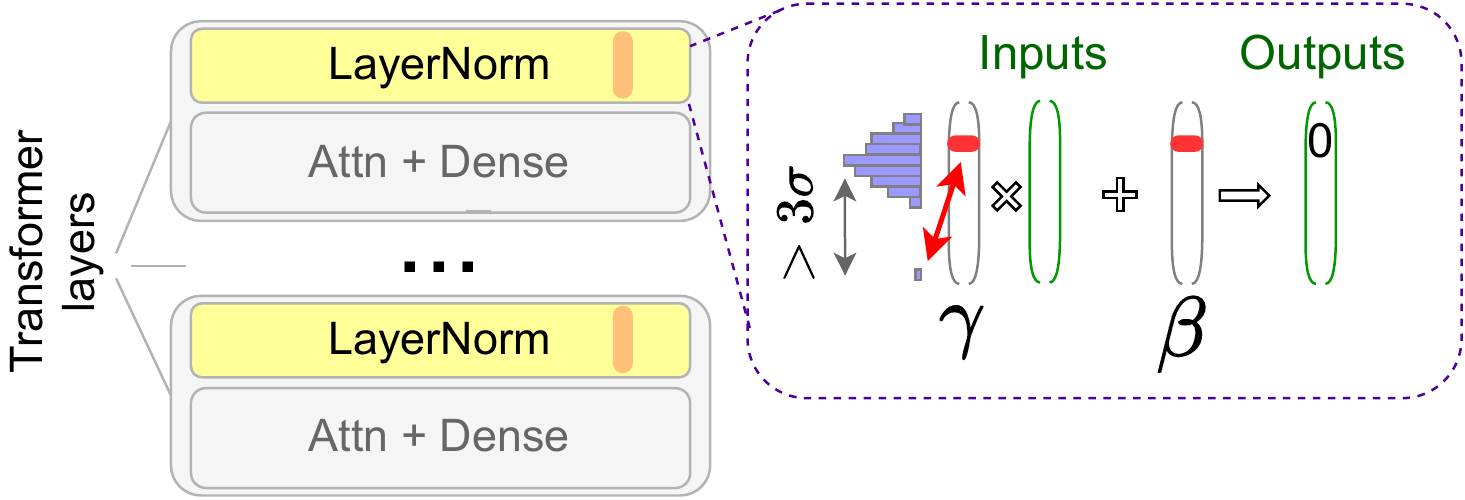}
    \caption{Simplified illustration of our approach to disabling LayerNorm weights. We consistently set the outlier weights $\gamma$ and $\beta$ of the output LayerNorm to zeroes, which results in masking of the corresponding features in the output vectors. We repeat the procedure for all of the Transformer layers of the encoder.}
    \label{fig:diagram}
\end{figure}

To identify the outlier weights in BERT-like models, we consider all the output components in each encoder layer. We compute the mean and standard deviation of the bias and scaling factors of the output LayerNorm. We identify the dimensions where both of these weights are at least 3$\sigma$ from the mean. \autoref{fig:diagram} illustrates this heuristic. Further, we select the dimensions where this is consistently the case for at least  half of the model layers. We refer to these dimensions as outliers.

The described heuristic was used to identify the outlier dimensions in four out of six BERT models we considered: BERT-base, BERT-medium, BERT-small and mBERT.\footnote{BERT-medium and BERT-small come from the official Google repository (\url{https://github.com/google-research/bert}), and the other models from the HuggingFace (\url{https://github.com/huggingface/transformers}). Interestingly, we discover that the checkpoints of the same BERT-base configuration provided by different repositories (Huggingface vs. Google) have the outliers in different locations; the outliers also have different values: positive vs. negative.}
For BERT-large, the deepest model we considered, the frequency constraint was relaxed to 1/3 of the layers. In RoBERTa \cite{liu2019roberta}, the distribution of the scaling factors was a little different from BERTs, and we relaxed the standard deviation constraint down to 2 sigmas to detect the outliers. In Section \ref{sec:outliers_all} of Appendix, we report positions of outlier weights identified for all models.

\subsection{Disabling}
\label{sec:disabling}

To quantify the effect of the outlier weights on BERT, we disable them 
and examine how this affects model performance.  We set the outlier weights to zeros across all layers 
and report model performance on a) masked language modeling and b) downstream GLUE tasks \cite{WangSinghEtAl_2018_GLUE_Multi-Task_Benchmark_and_Analysis_Platform_for_Natural_Language_Understanding}. 

Since different model components may affect performance, we also looked at all the parameter vectors and matrices in BERT that have the same dimensionality as the output embeddings. These included key, query and value transformations\footnote{To find outliers in weight matrices, we compute the L$_1$ norm for each row. The total number of rows is the same as the dimensionality of the layer output embedding (e.g. 768 for BERT-base). From this distribution of row-norms, we select those row indices for which the magnitude is {$3\sigma$} away from the mean of the distribution.}, output LayerNorm, attention LayerNorm, and input embedding layers. In order to examine the effect of these components on model performance, we masked the identified outlier weights of a given component simultaneously across all layers at the same dimensional position. When working with the matrices, we set to zeros the entire row of weights corresponding to the outlier positions across all Transformer layers. Similarly, for LayerNorm, we set the scaling factor and the bias at the outlier position across all Transformer layers to zero. In both cases, this results in ``masking'' of the output vector's feature at the specified dimension after a forward pass through that layer.

Although the same dimensions repeatedly show up as outliers across different model components, in the preliminary experiments, we found that disabling the weights of the input embedding layers and of the linear layers produced no significant change in performance or in the output embedding space, so we did not pursue this direction further. However, the outlier weights of the output LayerNorm had an unexpectedly large effect on the model, and this is what we focus on in most of our experiments.

\subsection{Visualization}

\begin{figure}[t]
    \centering
    \includegraphics[width=\linewidth]{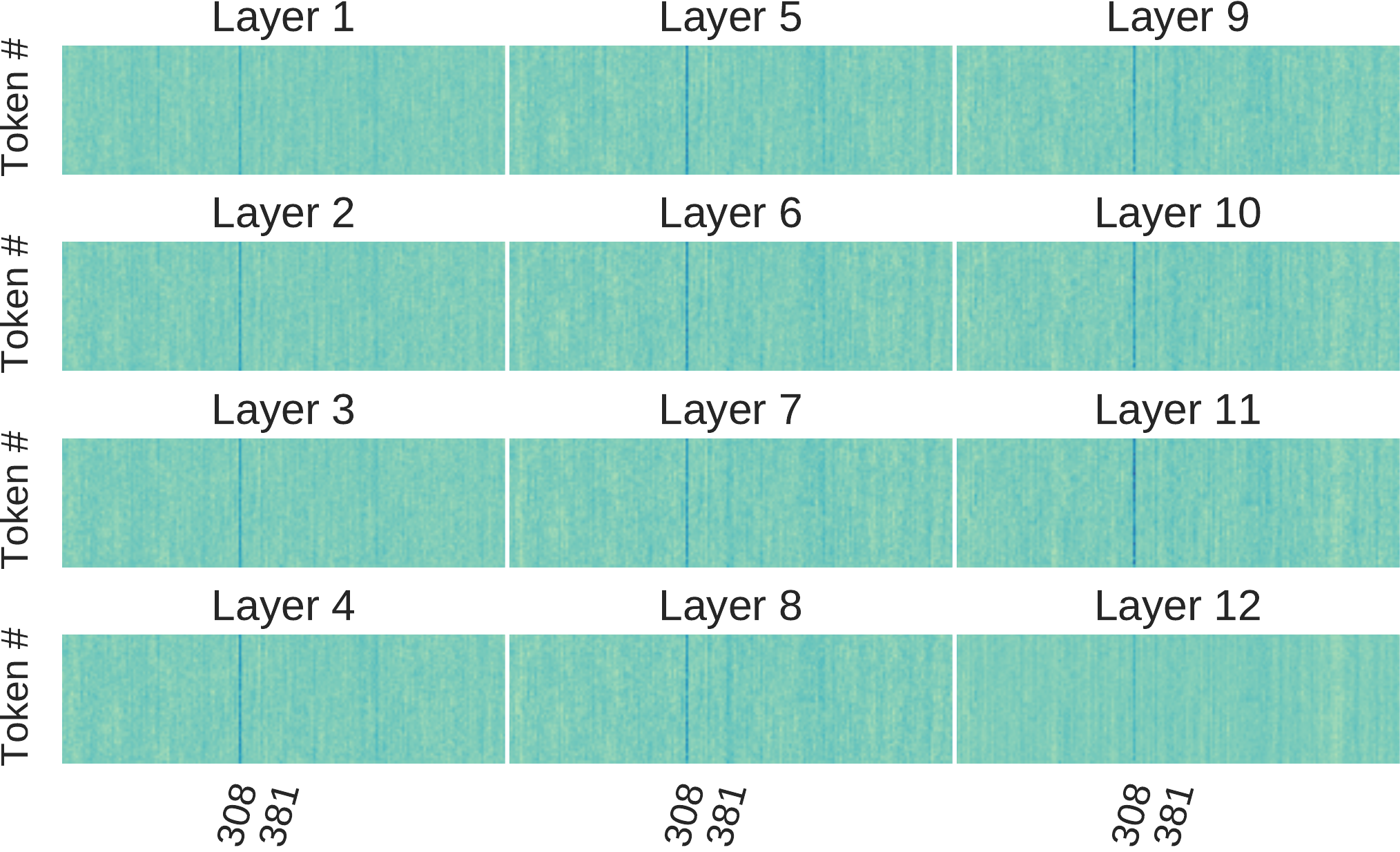}
    \caption{Outlier LayerNorm features \texttt{308}, \texttt{381} in BERT-base-uncased (randomly sampled input).}
    \label{fig:outliers}
\end{figure}

As an example, let us consider the two outlier dimensions that the above method identifies for BERT-base-uncased model: \texttt{308} \footnote{All dimensions in the paper are zero-indexed.} and \texttt{381}. 
\autoref{fig:outliers} shows a heatmap of the output embedding of each layer (one pixel per row) for a random passage from WikiText \cite{merity2016pointer}.  Since the output embeddings are produced by LayerNorm, the outlier dimensions with unusually high or low-magnitude weights should be visible in the heatmap.

As seen in \autoref{fig:outliers}, dimension \texttt{308} consistently produces high-magnitude weights in the output embeddings in most BERT layers; feature \texttt{381} shows visibly high values in layers 7-10. 
The magnitude of a given feature depends both on the LayerNorm scaling factor and the bias (\autoref{eq:scale_shift}). We find that both contribute to the outlier effect, to various degrees in different layers. 
See \autoref{tab:stats} in the Appendix for statistics on all scaling factors and biases in BERT-base.

Features that show high magnitude throughout the model are expected to distort the resulting embedding space.  In a brief experiment, we found that the vector representations of up to 95\% of the input tokens from the WikiText corpus have abnormally high magnitudes at the dimensions corresponding to the outlier weights we identified. 
We found that this embedding distortion is not attributable to the input embedding layer (Layer 0). We did this by manually setting select embedding weights of the three channels of the input embedding layer (token, token type, and position) to zero (along with the weights of the following normalization layer).

The fact that the embedding distribution we observed is not uniform is in line with observations by \citet{ethayarajh2019contextual} who concluded that BERT embeddings are highly anisotropic and form a cone-like shape in the hidden space. The outlier weights we identify are likely the cause of this, since after they are removed, the embedding space becomes relatively uniform.

\section{Effects of Disabling Outlier Weights} \label{sec:experiments}
In this section, we consider the effects of disabling outlier weights in BERT on language modeling (\cref{sec:mlm}) and on downstream tasks (\cref{sec:glue}). We also investigate whether other Transformers exhibit a similar phenomenon (\cref{sec:others}).

\begin{table}[t]
\footnotesize
\centering
\begin{tabular}{p{0.2cm}p{6cm}}
\rotatebox[origin=r]{90}{Input} & Ghostbusters was [\textcolor{teal}{released}] on June 8 , [\textcolor{teal}{1984}] , to critical [\textcolor{teal}{acclaim}] and became a cultural phenomenon . It was well [\textcolor{teal}{received}] for its deft blend of comedy, [\textcolor{teal}{action}] , and horror , and Murray ' s performance was [\textcolor{teal}{repeatedly}] singled out for praise . \\
\rotatebox[origin=r]{90}{RoBERTa} & Ghostbusters was [\textcolor{green}{released}] on June 8 , [\textcolor{red!75!green}{1986}] , to critical [\textcolor{green}{acclaim}] and became a cultural phenomenon . It was well [\textcolor{green}{received}] for its deft blend of comedy,  [\textcolor{green}{action}] , and horror , and Murray ' s performance was [\textcolor{red!75!green}{often}] singled out for praise . \\

\rotatebox[origin=r]{90}{Random} & Ghostbusters was [\textcolor{green}{released}] on June 8 , [\textcolor{red!75!green}{1986}] , to critical [\textcolor{green}{acclaim}] and became a cultural phenomenon . It was well [\textcolor{green}{received}] for its deft blend of comedy,  [\textcolor{green}{action}] , and horror , and Murray ' s performance was [\textcolor{red!75!green}{particularly}] singled out for praise . \\
\rotatebox[origin=r]{90}{Outliers}& \textcolor{red}{\{} \textcolor{red}{lock} was [\textcolor{red}{never}] on June 8 , [\textcolor{red}{</s>}] , to \textcolor{red}{rely} [\textcolor{red}{,}] and . It was well [\textcolor{red!75!green}{known}] for its \textcolor{red}{acker} of comedy , [\textcolor{red}{dinner}], and horror , and Murray ' s was [\textcolor{red}{ever}] \textcolor{red}{,} \textcolor{red}{</s>} \textcolor{red}{</s>} \textcolor{red}{)}
\end{tabular}
\caption{Input masked tokens (blue) are given in brackets. RoBERTa correctly reconstructs 4 out of 6 masked tokens (green), and fills in plausible (brown) predictions for the remaining 2 tokens. RoBERTa with 2 randomly disabled LayerNorm dimensions works almost the same as the base model. However, RoBERTa with 2 outlier LayerNorm dimensions makes incorrect and implausible (red) predictions, and changes the hidden token states significantly enough to map the unmasked input tokens to other, often non-sensical words. In this example, we do not show the special tokenizer tokens.}
\label{tab:sample_attack}
\end{table}

\begin{figure*}[ht]
    \centering
    \includegraphics[width=\linewidth]{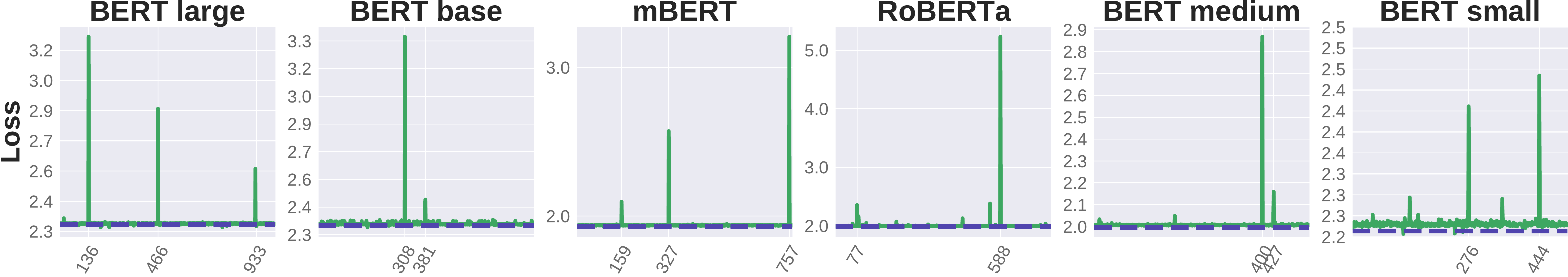}
    \caption{Language model cross-entropy loss (green) after 
    the scaling factor and bias for a given dimension are set to zero, one at a time. The dashed blue line shows the loss achieved by the full model.}
    \label{fig:lm_one_at_a_time}
\end{figure*}

\begin{table*}[ht]
\small
\centering
\begin{tabular}{p{0.2cm}p{1.5cm}|lrlrlrlrlrlr}
\toprule
& \multirow{2}{*}{} & \multicolumn{2}{c}{\textbf{BERT-large}} & \multicolumn{2}{c}{\textbf{BERT-base}} & \multicolumn{2}{c}{\textbf{mBERT}} & \multicolumn{2}{c}{\textbf{RoBERTa}} & \multicolumn{2}{c}{\textbf{BERT-medium}} & \multicolumn{2}{c}{\textbf{BERT-small}} \\
& & \#w & CE & \#w & CE & \#w & CE & \#w & CE & \#w & CE & \#w & CE \\
\midrule
& Baseline & 0 & 2.28 & 0 & 2.30 & 0 & 1.93 & 0 & 1.99 & 0 & 2.00 & 0 & 2.26 \\
\midrule
\multirow{2}{*}{\rotatebox[origin=c]{90}{{\scriptsize Single}}} &

Random & 48 & 2.29 & 24 & 2.31 & 24 & 1.95 & 24 & 2.03 & 16 & 2.01 & 8 & 2.26 \\
& Top outlier & 48 & 3.22 & 24 & 3.33 & 24 & 3.21 & 24 & 5.23 & 16 & 2.87 & 8 & 2.44 \\
\midrule
\multirow{2}{*}{\rotatebox[origin=c]{90}{{\scriptsize All}}} &  Random & 144 & 2.29 & 48 & 2.31 & 72 & 1.96 & 48 & 2.00 & 32 & 2.04 & 16 & 2.28 \\
& All outliers & 144 & \textbf{5.49} & 48 & \textbf{4.53} & 72 & \textbf{6.92} & 48 & \textbf{7.85} & 32 & \textbf{3.21} & 16 & \textbf{2.93} \\
\bottomrule
\end{tabular}
\caption{Cross-entropy (\textit{CE}) on the validation set of WikiText when the LayerNorm weights are zeroed. \textit{Single}  shows the performance when the most damaging outlier is disabled vs. disabling one non-outlier feature
(averaged over all non-outliers disabled one at a time). \textit{All} shows performance for when all outliers in a given model are disabled vs. disabling an equal number of randomly selected non-outliers (averaged over 1000 runs). \textit{\#w} indicates the total number of modified weights in a given model. 
}
\label{tab:lm_all}
\end{table*}

\subsection{Masked Language Modeling} \label{sec:mlm}

Our key finding is that disabling the outlier dimensions significantly degrades the quality of the language model, even though fewer than 0.001\% weights of the model are affected. \autoref{tab:sample_attack} shows a sample output before and after the LayerNorm outliers are disabled in RoBERTa\footnote{More sample outputs are given in the Appendix.}. It is clear that the quality of the language model degrades dramatically, while disabling an equal number of randomly selected non-outliers has almost no effect.

To quantify this effect, we measure the cross-entropy loss before and after disabling each dimension of the LayerNorm on at a time. We do this on the validation subset of the WikiText corpus \cite{merity2016pointer}. We use the standard maximum sequence length of 256 and the token masking probability of 0.15.

All tested models show surprising sensitivity to zeroing out the  weights at the outlier positions across all layers of the model  (\autoref{fig:lm_one_at_a_time}). For example, removing only 24 parameters (the scaling factor and the bias of a specific LayerNorm dimension across all 12 layers) increases RoBERTa's loss by almost a factor of 4.

\autoref{tab:lm_all} shows even more drastic effects when scaling factors and biases for all the outlier dimensions are disabled simultaneously. 
For comparison, we randomly sample an equal number of non-outlier dimensions and disable the corresponding LayerNorm weights throughout the model. We report the loss averaged over 1000 runs.

\begin{figure*}[ht]
    \centering
    \includegraphics[width=\linewidth]{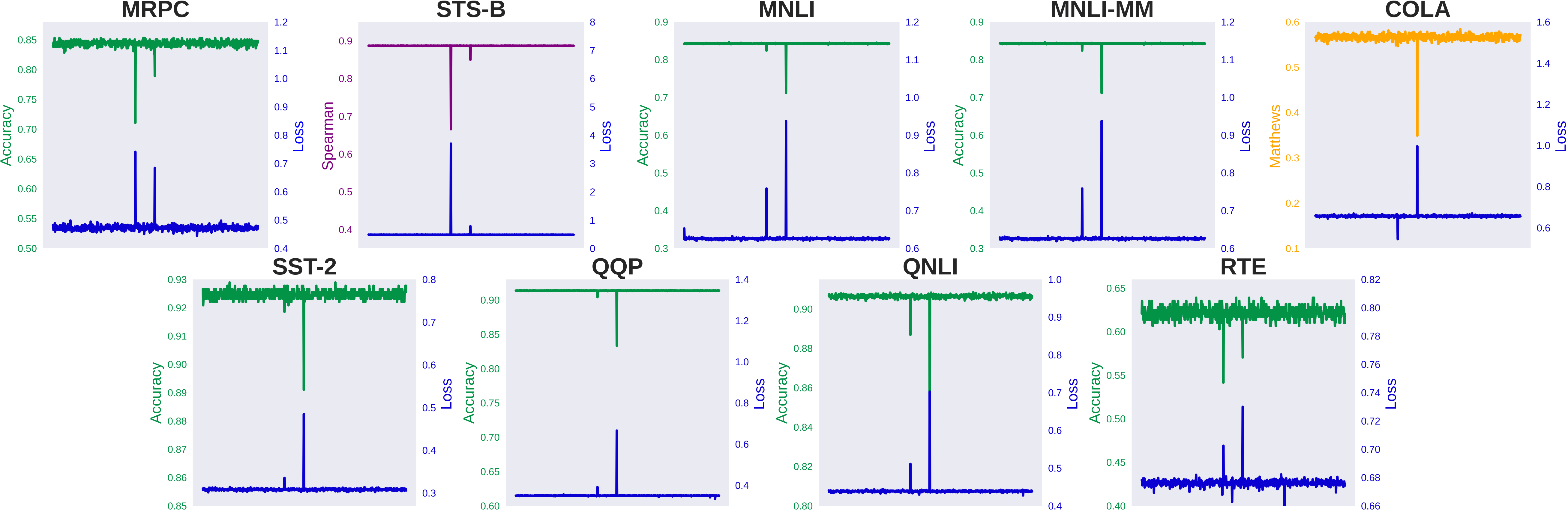}
    \caption{Performance of BERT-base on GLUE tasks when one LayerNorm weight at a time is disabled throughout the model. Dimensions are shown on the X axis. Loss (blue) and accuracy, or correlation coefficients, where applicable (other colors)  are shown.}
    \label{fig:donwstream_one_by_one}
\end{figure*}

\begin{table*}[t]
    \small
    \centering
    \begin{tabular}{p{0.2cm}lrrrrrrrrr}
        \toprule
        & & \textbf{MRPC} & \textbf{STS-B} & \textbf{MNLI} & \textbf{MNLI-mm} & \textbf{COLA} & \textbf{SST-2} & \textbf{QQP} & \textbf{QNLI} & \textbf{RTE} \\
        \midrule
         
        & Baseline (full model) & 87.2 & 88.8 & 84.1 & 84.2 & 56.8 & 92.5 & 89.8 & 90.6 & 61.7 \\
        \midrule
        
        \multirow{5}{*}{\rotatebox[origin=c]{90}{Post-ft}} &
        Non-outlier$^\dag$ & +0.3 & -0.1 & -0.2 & -0.1 & +0.2 & 0 & -0.1 & 0 &  -0.4\\
        & Outlier-\texttt{308} & -10.5 & -23.4 & -2.2 & -1.8 & -2.16 & -0.6 & -1.0 & -1.9 & -7.2 \\
        & Outlier-\texttt{381} & -4.6 & -4.4 & -13.7 & -13.0 & -22.2 & -3.4 & -10.8 & -7.3 & -5.0 \\
        
        \cmidrule{2-11}
        
        & Random non-outlier pair$^\ddag$ & -1.1 & 0.0 & +0.3 & +0.2 & -0.5 & +0.1 & +0.1 & 0 & +0.5\\
        & Outliers \texttt{308} + \texttt{381} & \textbf{-8.6} & \textbf{-44.1} & \textbf{-27.9} & \textbf{-27.2} & \textbf{-32.3} & \textbf{-20.8} & \textbf{-13.0} & \textbf{-12.2} & \textbf{-10.0}  \\
        \midrule
        \multirow{4}{*}{\rotatebox[origin=c]{90}{Pre-ft}} 
        & Random non-outliers$^*$ & -0.3 & -0.05 & -0.2 & -0.2 & +0.9 & -0.06 & -0.2 & -0.3 & +0.6 \\
        & Outlier-\texttt{308} & +0.3 & -0.9 & -0.5 & +1.7 & -0.3 & +0.7 & -0.1 & 0 & -5.1 \\
        & Outlier-\texttt{381} & -2.4 & -0.7 & -0.6 & -0.5 & -0.9 & -1.2 & -0.7 & -1.4 & +4.0 \\
        \cmidrule{2-11}
        & Outliers \texttt{308} + \texttt{381} & -1.1 & -1.6 & -1.4 & -0.7 & -2.9 & -1.7 & -0.7 & -2.3 & -0.7   \\
        \bottomrule
    \end{tabular}
    \caption{Performance of the pretrained BERT-base model vs. different configurations with disabled outlier weights: post fine-tuning (\textit{post-ft}) and pre-fine-tuning (\textit{pre-ft}). $^\dag$We disable each of non-outlier dimension parameters one at a time and average over them. $^\ddag$For the random sampling of the pairs of non-outlier dimensions, we report averages over 1000 runs. $^*$For the pre-finetuning experiment where random non-outlier parameters are disabled, we sample 10 non-outlier dimensions randomly and disable LayerNorm weights and biases for them across the entire model.}
    \label{tab:downstream_all}
\end{table*}

\begin{figure*}[ht]
    \centering
    \includegraphics[width=\linewidth]{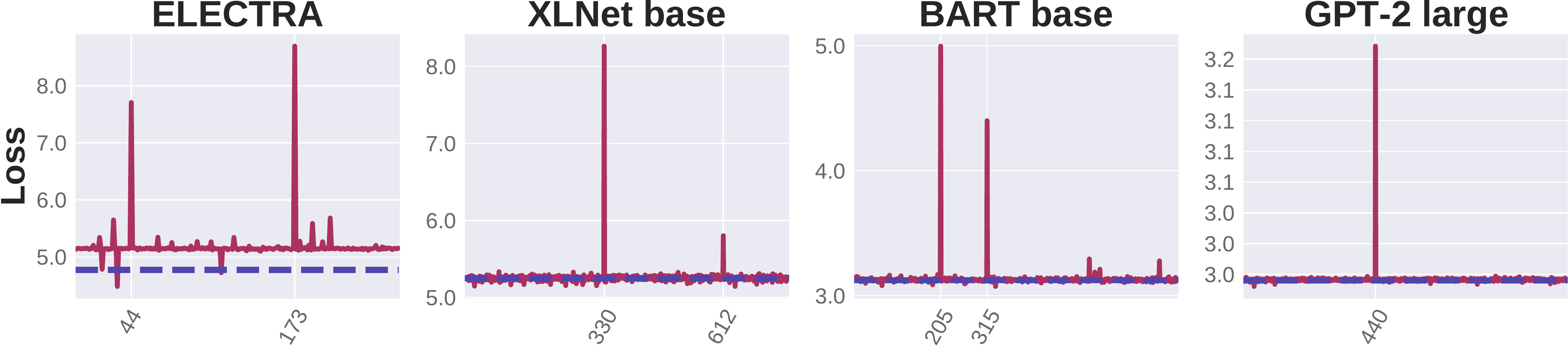}
    \caption{Performance (loss) of other Transformer models after a given LayerNorm scaling factor and bias are set to zero, one at a time. For GPT-2, the dense layer weights and biases are disabled, as it is an instance of a pre-LN model. The dashed blue line shows the loss for the full model.}
    \label{fig:generative}
\end{figure*}

\begin{table*}[ht]
    \centering
    \footnotesize
    \begin{tabular}{lrrrrrrr}
    \toprule
    \multirow{2}{*}{\textbf{Model}} & \multirow{2}{*}{\textbf{Task}} & \multirow{2}{*}{\textbf{Data}} & \multirow{2}{*}{\textbf{\#Params}} & \multicolumn{2}{c}{\textbf{Disabled}} & \multicolumn{2}{c}{\textbf{Performance}} \\
    & & & & \# dims & weight \% & Baseline & Disabled\\
    \midrule
    ELECTRA & MLM & WikiText & 110M & 2& $4\times10^{-5}$ & 4.8 & \textbf{8.1}\\
    XLNet & PLM & WikiText & 120M & 2 & $4\times10^{-5}$ & 5.2 & \textbf{8.4}\\
    BART & Summ. & CNN/Daily Mail & 140M & 2 & $9\times10^{-6}$ & 3.1 & \textbf{4.4} \\
    GPT2 & CLM & WikiText & 770M & 1 & $0.024$ & 3.0 & \textbf{3.2}\\
    \bottomrule
    \end{tabular}
    \caption{Loss increase in other Transformer models after layer normalization scaling factors and biases highlighted in \autoref{fig:generative} are set to zero, compared to the baseline configuration. For GPT-2, the dense layer weights and biases are disabled instead, since it is a pre-LN model. \textit{MLM}, \textit{PLM}, and \textit{CLM} stand for masked, permutation, and causal language modeling objectives, respectively. \textit{Summ.} stands for the summarization task. \textit{\#dims} denotes the total number of dimensions modified. }
    \label{tab:generative}
\end{table*}

\subsection{BERT Downstream Tasks} \label{sec:glue}

In order to investigate the effect of outlier weights on downstream performance, we evaluate BERT-base on the GLUE benchmark tasks \cite{WangSinghEtAl_2018_GLUE_Multi-Task_Benchmark_and_Analysis_Platform_for_Natural_Language_Understanding}, with the exclusion of Winograd Schema Challenge, which BERT generally fails to learn \cite{PrasannaRogersEtAl_2020_When_BERT_Plays_Lottery_All_Tickets_Are_Winning}. We use the evaluation split of the GLUE benchmark for which the labels are publicly available. As described above, BERT has two outlier dimensions, \texttt{308} and \texttt{381}. 
We consider the following two sets of experiments:
\paragraph{1. Disable post fine-tuning.}
    We fine-tune BERT on every GLUE task, then disable the outlier LayerNorm parameters (scaling factor and bias pair) across all layers as described in \autoref{sec:disabling}. We experiment with disabling each of the two detected outliers both individually and simultaneously in pre-trained BERT-base. We compare the resulting performance to (a) removing the LayerNorm weights for all other hidden dimensions one-by-one, and (b) randomly sampling pairs of non-outlier dimensions so as to disable them simultaneously.
\paragraph{2. Disable pre-fine-tuning.} We disable the layer norm parameters for the outlier dimensions \textit{prior} to fine-tuning. Our goal is to check whether the fine-tuning allows the transformers to recover the information from rest of the parameters. 

For all the fine-tuning runs, we set the learning rate to 5e-5, batch size to 64 and train for 4 epochs across all the experiments. 
Since BERT performance varies a lot due to task-specific initialization \cite{DodgeIlharcoEtAl_2020_Fine-Tuning_Pretrained_Language_Models_Weight_Initializations_Data_Orders_and_Early_Stopping}, we use the same initialization across all experiments.

\autoref{fig:donwstream_one_by_one} shows model performance when LayerNorm outliers are disabled one at a time. 
\autoref{tab:downstream_all} compares task-specific performance of the outlier-disabled and the full model for each task. 

The main takeaway from the \textit{post fine-tuning} experiments is that disabling one or the other of the outlier dimensions (or both) drastically degrades model performance on downstream tasks.  Which of them affect downstream performance the most is highly task-dependent.  For example, the outlier dimension \texttt{308} has little effect on CoLA(-2.16) but a large effect on STS-B(-23.4), and for \texttt{381} it's the opposite. On SST-2, QNLI, RTE neither outlier drops the performance by over 10 points individually, but disabling them both has a strong adverse effect.

In general, disabling two outliers together causes more severe damage across the board than disabling a single outlier. The overall performance drop is task-specific. The most adversely affected tasks are STS-B (-44.1) and CoLA (-32.3), which are the regression tasks that also suffered the most in pruning experiments by \citet{PrasannaRogersEtAl_2020_When_BERT_Plays_Lottery_All_Tickets_Are_Winning}. However, MNLI and SST-2 are classification tasks, and both of them also lose over 20 points.
Note that disabling random non-outlier dimensions (either alone, or in pairs) has negligible effect on performance across tasks. Disabling one outlier and one random non-outlier has the same effect as disabling a single outlier.

In \textit{pre-fine-tuning} experiments, the question we ask is whether the model can recover from the handicap we introduce, and still learn the task. We expect it to mostly recover, since even randomly initialized BERT without any pre-training can be fine-tuned to solve GLUE tasks fairly well \cite{KovalevaRomanovEtAl_2019_Revealing_Dark_Secrets_of_BERT}. In this case, since most of the pretrained sub-networks are still accessible to the classifier, we expect to see a strong recovery close to the baseline. We find this to be the case, however, the model is more adversely affected: -1.1 on average when two outliers are disabled, compared to disabling 10 non-outliers (-0.31 on average).

\subsection{Outliers in Other Transformers}
\label{sec:others}

Above, we described our methodology for identifying outliers in BERT models and studied how they affect model performance. 
In this section, we show that a similar phenomenon is observed in other popular Transformers: BART-base, ELECTRA-base generator, XLNet-base, and GPT-2 large.

Since our goal here was merely to confirm the existence of outlier dimensions, in these experiments, we simply disabled individual LayerNorm dimensions of the encoder part of the model across all Transformer layers. For GPT-2, we modify the weights of the dense output layer instead of the LayerNorm weights, since it uses the pre-LayerNorm (pre-LN) configuration (i.e., LayerNorm is placed before the output feature-producing feed-forward layer).

We perform this experiment for all models, measuring how this affects the loss function in the native pre-training tasks of three models: permutation language modeling task for XLNet, causal language modeling for GPT-2, and masked language modeling for the generator component of ELECTRA. For BART, we found that the pre-trained model had unusually high perplexity out of the box, and we substituted the modeling task with summarization on the CNN/Daily Mail dataset \cite{hermann2015teaching}.  

As with BERT, our results (\autoref{fig:generative}) suggest that for each model, there are a few distinct dimensions which disrupt performance significantly more than the rest.
We identify a few most impactful dimensions and also disable them at once, as reported in \autoref{tab:generative}. The effect on perplexity is the least pronounced for GPT2, which we attribute to the fact that the model is significantly larger than the others, which may make it more robust to the disabling of individual weights. However, we found that when six dimensions are disabled simultaneously, the perplexity increases by over 300 times.

\section{What Makes Outlier Weights Special}
\label{sec:analysis}

\subsection{Magnitude or Location?} \label{sec:ablation}
In this section, we conduct two experiments to validate our proposed criteria for selecting BERT weights to be disabled.  Specifically, we want to understand if the same effects would be observed (1) with magnitude-based selection of LayerNorm parameters to be disabled, or (2) using our selection method, but disabling the outlier dimensions only in the first (input) layer or in the later layers (which may have a more direct effect on the output).  We use the drop in the loss value for this comparison.

First, we compare the effect of disabling the selected dimensions (\texttt{308} and \texttt{381} in BERT-base, disabled individually or together -- i.e. disabling either 12 or 24 LayerNorm scaling factor and bias pairs) to disabling of the following alternatives:

\begin{itemize*}
\item \textbf{Random}: disable 12 or 24 randomly selected pairs of LayerNorm scaling factor and bias pairs in the entire model;
\item \textbf{Largest Scaling Factor (LSF)}: sort the LayerNorm scaling factors in the model by magnitude and disable the top 12 (or 24) scaling factors and the corresponding biases;
\item \textbf{Largest Bias (LB)}: repeat the above using the LayerNorm biases instead, i.e. select the top 12 (or 24) LayerNorm biases and disable the corresponding scaling factor / bias pairs.
\end{itemize*}

\noindent
\autoref{tab:ablation_magnitude} suggests that simple magnitude-based pruning of the output LayerNorm results in a much smaller degradation. As compared to disabling both BERT outliers, the magnitude-driven pruning of the same number of weights results in the value of cross-entropy that is 1.3x smaller.

\begin{table}[t]
    \centering
    \footnotesize
    \begin{tabular}{lrr|rr}
    \toprule
     & \textbf{CE} & \multicolumn{1}{r}{\textbf{\#w}} & \multicolumn{1}{r}{\textbf{CE}} & \textbf{\#w} \\
    \midrule
    baseline (full model) & 2.30 & 0 & 2.30 & 0 \\
    \midrule
    Random & 2.31 & 24 & 2.32 & 48 \\
    LSF & 2.72 & 24 & 2.74 & 48 \\
    LB & 3.21 & 24 & 3.42 & 48 \\
    Outlier-\texttt{308} & 3.32 & 24 & \multirow{2}{*}{$\Big\}$\textbf{4.53}} & \multirow{2}{*}{48} \\
    Outlier-\texttt{381} & 2.44 & 24 &  &  \\
    \bottomrule
    \end{tabular}
    \caption{BERT-base cross-entropy loss (\textit{CE}) on the WikiText validation data when one (left) or two (right) outlier dimensions are disabled at a time, compared to magnitude-based pruning approaches (LSF and LSB). \textit{\#w} denotes the total number of modified weights.}
    \label{tab:ablation_magnitude}
\end{table}

Looking at the effects of disabling the outlier dimensions only in a subset of layers, \autoref{tab:ablation_layers} shows that modifications made to the input layer have little to no effect. Interestingly, switching off the weights in the last Transformer layer, which is used for computing inputs for task-specific classifiers, also does not disrupt the model. However, as we begin to disable earlier layers and the number of layers with disabled weights increases, we observe progressively larger loss values.

\begin{table}[!h]
\small
\centering
\begin{tabular}{lrrrrrr}
\toprule
 & \multicolumn{1}{c}{\textbf{1}} & \multicolumn{1}{c}{\textbf{12}} & \multicolumn{1}{c}{\textbf{11--12}} & \multicolumn{1}{c}{\textbf{9--12}} & \multicolumn{1}{c}{\textbf{7--12}} & \multicolumn{1}{c}{\textbf{1--12}} \\
\midrule
CE & 2.33 & 2.40 & 2.67 & 2.81 & 2.87 & \textbf{4.53} \\
\#w & 4 & 4 & 8 & 16 & 24 & 48 \\
\bottomrule
\end{tabular}
\caption{BERT-base language modeling cross-entropy loss (\textit{CE}) on the validation set of WikiText corpus, shown by location and number of modified Transformer layers. \textit{\#w} denotes the total number of modified weights. }
\label{tab:ablation_layers}
\end{table}

We conclude that both the magnitude and the consistent emergence of outlier weights in the same locations across the model are responsible for the emergence of distinct embedding features that BERT heavily relies on.

\begin{figure*}[ht]
    \includegraphics[width=\linewidth]{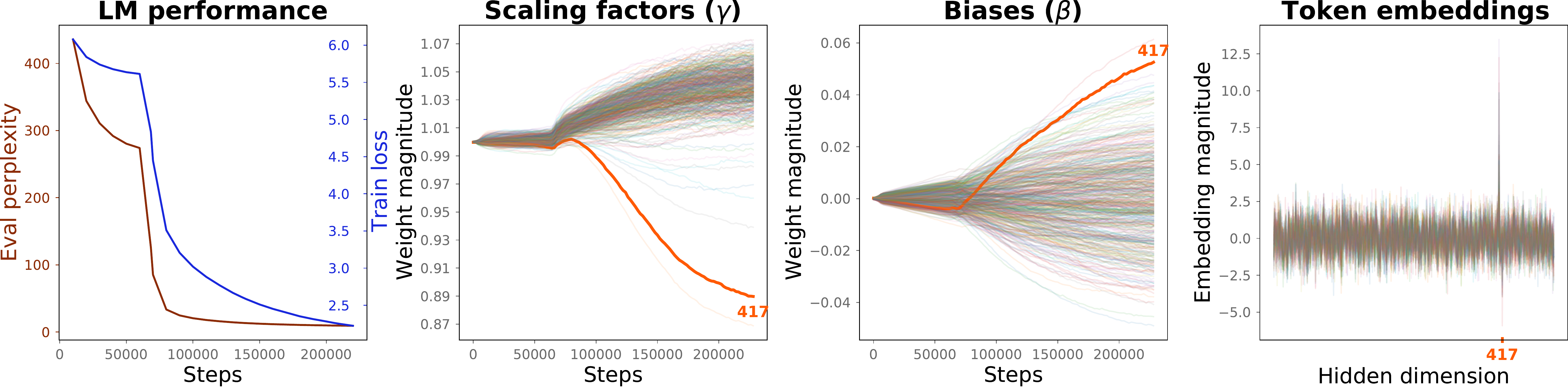}
        \caption{BERT-medium pre-training on the BookCorpus dataset. \textit{(left)} Evaluation perplexity (brown) and train loss (blue) as the training progresses. \textit{(middle)} The changes in the scaling factors and the biases of the output normalization layer. Each line corresponds to one of the 512 dimensions. We highlight (in orange) the \texttt{417}-th dimension, for which both the scaling factor and the bias fall out of the three sigma range at the end of pretraining. \textit{(right)} Token embeddings computed for an input sequence that was randomly sampled from the data. Each line corresponds to one input token. The outlier embedding values are marked at the same \texttt{417}-th dimension. All the plots are presented for the middle Transformer layer (4).}
    \label{fig:pretraining}
\end{figure*}

\subsection{How Do Outlier Weights Emerge?} \label{sec:pretraining}

In this section, we examine the emergence of outlier dimensions during pre-training.
To the best of our knowledge, there are no publicly available BERT pre-training checkpoints available to study these effects. We pre-train\footnote{We used two RTX-3090 GPUs for pre-training.} a BERT-medium model (chosen due to computational constraints) from scratch on the BookCorpus data \cite{zhu2015aligning} and track statistics of the LayerNorm scaling factors and biases. We start from a randomly initialized BERT-medium configuration that has 8 layers with the hidden dimensionality of 512 units. We save checkpoints of the model every 2000 steps, and we track the output LayerNorm weights across all of the model's layers as the training progresses.

\autoref{fig:pretraining} shows that both scaling factors and biases begin to diverge from their initialization values quite early (after approximately 50k steps) in the training process. At roughly the same point, both training loss and evaluation perplexity begin to fall off. An interesting question for future work is to clarify whether there is a causal relationship here.

Although the published BERT-medium model had two outlier dimensions, our model had only one dimension for which \textit{both} the scaling factor and the bias exceed their corresponding means by more than three standard deviations. 

Due to the gradual emergence of these outliers during pre-training, there is a possibility that thresholding their distance from the rest of the parameters can be used as a litmus test for when the pre-training is complete.  We leave the investigation of this hypothesis to future work.

\section{Implications and Future Work}

The outlier dimension effect we identified may enable \textbf{attacks on Transformer-based encoders} that could be used to degrade the model quality while modifying very few weights. Further, since these perturbations to the model do not cause the model to break completely, this may lead to late detection of this attack. To curtail the risks of this attack from affecting deployed Transformer models, we would suggest simple measures such as storing the file checksums for the trained models at a secure location and verifying that the deployed model file matches the checksums.

Another direction for exploiting the phenomenon of outlier dimensions is \textbf{pruning}. The studies of model compression using unstructured pruning typically do not consider whether the pruned weights were in the same position throughout the model. Our work suggests that if the outliers were disabled consistently, the drop in performance could be expected to be larger than for random or magnitude-based pruning.

Finally, future work could consider outlier dimensions in the context of \textbf{weight initialization}. Our experiments suggest that these dimensions are a normal emergent property of Transformer pre-training. It is possible that higher performance or faster convergence could be achieved by manipulating the initialization to encourage such outliers and experimenting with their number. 

\section{Conclusion}

The main contribution of our work is isolating the phenomenon of a small number of outlier dimensions in Transformer layer outputs which significantly disrupt performance while modifying less than 0.0001\% of all parameters of the model. We attribute this phenomenon to an interaction of high-magnitude scaling factors and biases in the same dimension throughout the model, rather than magnitude alone. It emerges early in the training and consistently warps the embedding space.

In case of BERT, the layer output component is LayerNorm. We introduce a method to isolate these outlier dimensions for BERT, and we show that the phenomenon is present in six models of BERT family that we examine. It is also present in four other Transformer-based models (ELECTRA, XLNet, BART, and GPT-2), although the effect of their disabling varies.

\section{Acknowledgments}

We would like to thank the anonymous reviewers for their insightful comments. This  work  is  funded  in  part  by the  NSF  award  number  IIS-1844740  to  Anna Rumshisky.

We would like to note that 
in an independent study concurrent to ours, \citet{luo2021positional} report a similar effect of outlier values in the embedding space of BERT and RoBERTa. 
We invite the reader to look into their study for more details on the outlier features from the perspective of word embeddings.

\section{Impact statement}

This work analyzes the behavior of layer normalization in the popular BERT family of models, using standard benchmarks. No new models are presented, and no data was collected.

We estimate that experiments presented in this paper consumed 436.8 kWh of energy which resulted in a 108 Kilograms of CO2 emissions computed using the regional average of power generation to carbon emissions.  

\bibliography{acl2021}
\bibliographystyle{acl_natbib}

\clearpage

\onecolumn
\section{Appendix}

\subsection{Candidate outlier dimensions of the models.} \label{sec:outliers_all}
For each of the BERT-like models we experimented with we present the outlier candidate weights, detected as described in \autoref{sec:methodology}.
\begin{table*}[ht]
\small
\centering
\begin{tabular}{p{5cm}p{7cm}}
\toprule
\textbf{Model component} & \textbf{Outliers} \\
\midrule
output.dense.weight &	275, 276, 444 \\
attention.output.dense.bias &	193 \\
attention.output.LayerNorm.weight  &	275, 276, 444 \\
attention.output.LayerNorm.bias &	276, 444 \\
output.LayerNorm.weight	& 121, 262, 444, 276 \\
output.LayerNorm.bias	& 276, 444 \\
\bottomrule
\end{tabular}
\caption{BERT-small outlier dimension candidates across model components.}
\end{table*}

\begin{table*}[ht]
\small
\centering
\begin{tabular}{p{5cm}p{7cm}}
\toprule
\textbf{Model component} & \textbf{Outliers} \\
\midrule
attention.output.dense.bias &	92, 400, 476, 17 \\
output.dense.weight & 	400 \\
output.dense.bias & 400 \\
attention.output.LayerNorm.weight & 17, 400, 430 \\
attention.output.LayerNorm.bias & 192, 400 \\
output.LayerNorm.weight & 11, 193, 393, 427, 400 \\
output.LayerNorm.bias	& 400, 427 \\

\bottomrule
\end{tabular}
\caption{BERT-medium outlier dimension candidates across model components.}
\end{table*}

\begin{table*}[ht]
\small
\centering
\begin{tabular}{p{5cm}p{7cm}}
\toprule
\textbf{Model component} & \textbf{Outliers} \\
\midrule
output.dense.weight & 308, 381 \\
output.dense.bias & 308 \\
attention.output.dense.bias & 308 \\
attention.output.LayerNorm.weight & 308, 381 \\
attention.output.LayerNorm.bias & 145, 308, 381 \\
output.LayerNorm.weight & 92, 145, 308, 381, 225 \\
output.LayerNorm.bias & 308, 381 \\
\bottomrule
\end{tabular}
\caption{BERT-base outlier dimension candidates across model components. }
\end{table*}

\begin{table*}[ht]
\small
\centering
\begin{tabular}{p{5cm}p{7cm}}
\toprule
\textbf{Model component} & \textbf{Outliers} \\
\midrule
attention.output.dense.bias & 757, 327 \\
output.dense.weight & 757, 159 \\
output.dense.bias & 159, 757 \\
attention.output.LayerNorm.weight & 159, 757, 327 \\
attention.output.LayerNorm.bias & 159, 757, 327 \\
output.LayerNorm.weight & 159, 757, 327 \\
output.LayerNorm.bias & 159, 757, 327 \\
\bottomrule
\end{tabular}
\caption{Multilingual BERT (mBERT) outlier dimension candidates across model components.}
\end{table*}

\begin{table*}[ht]
\small
\centering
\begin{tabular}{p{5cm}p{7cm}}
\toprule
\textbf{Model component} & \textbf{Outliers} \\
\midrule
output.dense.weight & 588 \\
output.dense.bias & 588, 494 \\
attention.output.dense.bias & 588 \\
attention.output.LayerNorm.bias & 77, 217, 453, 551, 588, 496, 731, 494 \\
output.LayerNorm.bias & 77, 453, 551, 588, 217, 240, 496, 61, 494 \\
\bottomrule
\end{tabular}
\caption{Base RoBERTa outlier dimensions across model components.}
\end{table*}

\begin{table*}[ht]
\small
\centering
\begin{tabular}{p{5cm}p{7cm}}
\toprule
\textbf{Model component} & \textbf{Outliers} \\
\midrule
attention.output.dense.bias & 466, 18 \\
output.dense.bias & 466, 750, 18, 933 \\
attention.output.LayerNorm.weight & 234, 466, 933 \\
attention.output.LayerNorm.bias & 9, 71, 136, 234, 327, 706, 466, 474, 929, 933, 18, 143 \\
output.LayerNorm.weight & 80, 136, 232, 234, 331, 466, 639, 665, 702, 724, 750, 763, 968, 315, 428, 933, 18, 506, 314 \\
output.LayerNorm.bias & 136, 466, 706, 327, 9, 929, 18, 143, 933 \\
\bottomrule
\end{tabular}
\caption{BERT-large outlier dimension candidates across model components.}
\end{table*}

\subsection{Scaling factor and bias statistics for BERT-base.}
For the BERT-base configuration, we present the detailed statistics on per-layer scaling factors and biases of the output LayerNorm (see \autoref{tab:stats}). We report per-layer means, standard deviations and counts of the weights falling out of the three sigma range. We also show the values of the outlier weights (\texttt{308} and \texttt{381}) along with their ranks, where the ranks are computed for the corresponding sorted arrays of weight magnitudes. Note that the outlier weights consistently appear to be \textit{among} the top largest or top smallest LayerNorm weights throughout the model, but are not necessarily the top-1 largest/smallest values.
 \begin{table*}[ht]
 \small
\begin{tabular}{p{1cm}lrp{1.4cm}p{1.4cm}|lrp{1.5cm}p{1.5cm}}
\toprule
\multicolumn{1}{l}{} & \multicolumn{4}{c}{\textbf{Scaling factors}} & \multicolumn{4}{c}{\textbf{Biases}} \\
\midrule
Transf. layer & mean / std & \multicolumn{1}{l}{\#$>3\sigma$} & 308 value / rank & 381 value / rank & mean / std & \multicolumn{1}{l}{\#$>3\sigma$} & 308 value / rank & 381 value/rank \\
\midrule
1 & 0.756 / 0.056 & 12 & 0.343 / 764 & 0.404 / 762 & -0.037 / 0.099 & 6 & -1.325 / 0 & 0.144 / 78 \\
2 & 0.870 / 0.069 & 24 & 0.400 / 765 & 0.374 / 766 & -0.034 / 0.086 & 8 & -0.678 / 0 & 0.277 / 5 \\
3 & 0.851 / 0.052 & 16 & 0.408 / 767 & 0.549 / 765 & -0.031 / 0.075 & 4 & -0.070 / 298 & 0.118 / 103 \\
4 & 0.811 / 0.044 & 11 & 0.562 / 764 & 0.388 / 767 & -0.033 / 0.052 & 7 & 0.075 / 174 & 0.114 / 50 \\
5 & 0.840 / 0.045 & 8 & 0.615 / 763 & 0.360 / 767 & -0.031/ 0.051 & 8 & 0.200 / 3 & -0.083 / 113 \\
6 & 0.832 / 0.037 & 7 & 0.692 / 763 & 0.411 / 767 & -0.032 / 0.060 & 6 & 0.403 / 0 & -0.394 / 1 \\
7 & 0.834 / 0.037 & 4 & 0.752 / 752 & 0.375 / 767 & -0.033 / 0.063 & 5 & 0.785 / 0 & -0.337 / 1 \\
8 & 0.810 / 0.030 & 4 & 1.163 / 0 & 0.335 / 767 & -0.033 / 0.065 & 2 & 0.959 / 0 & 0.304 / 1 \\
9 & 0.831 / 0.042 & 6 & 1.618 / 0 & 0.262 / 767 & -0.035 / 0.062 & 2 & 0.129 / 38 & 0.695/0 \\
10 & 0.801 / 0.060 & 7 & 1.437 / 0 & 0.254 / 764 & -0.032 / 0.057 & 9 & -0.415 / 2 & 0.258/4 \\
11 & 0.817 / 0.062 & 9 & 1.671 / 0 & 0.185 / 765 & -0.040 / 0.068 & 5 & -0.667 / 1 & 1.234/0 \\
12 & 0.633 / 0.027 & 13 & 0.273 / 767 & 0.536 / 758 & -0.019 / 0.050 & 5 & 0.225 / 0 & -0.021/531 \\
\bottomrule
\end{tabular}
\caption{The statistics of output LayerNorm weights (scaling factors and biases) for all of the Transformer layers of BERT-base.}
\label{tab:stats}
\end{table*}

\subsection{Sample language model outputs after disabling outlier LayerNorm weights.}
For RoBERTa and BERT, we randomly sample a set of sentences from  Wikipedia and BookCorpus, mask multiple input tokens, and use the models for token prediction. We compare the baseline (full) models with the models where select LayerNorm weights are zeroed out across all of the Transformer layers. In particular, we compare the setups where the outlier dimensions (two per model) are disabled as opposed to random dimensions (two per model).

\begin{table*}[]
\footnotesize
\centering
\begin{tabular}{p{0.2cm}p{4.5cm}p{4.5cm}p{4.5cm}}
\toprule
\rotatebox[origin=r]{90}{Input} &  Ghostbusters was [\textcolor{teal}{released}] on June 8 , [\textcolor{teal}{1984}] , to critical [\textcolor{teal}{acclaim}] and became a cultural phenomenon . It was well [\textcolor{teal}{received}] for its deft blend of comedy, [\textcolor{teal}{action}] , and horror , and Murray ' s performance was [\textcolor{teal}{repeatedly}] singled out for praise .
& a filmy coating of [\textcolor{teal}{dust}] and pebbles had settled onto the block , and [\textcolor{teal}{sami}] 's hand instinctively jerked forward to swipe the [\textcolor{teal}{scratchy}] debris off his cheek , then pulled [\textcolor{teal}{up}] short against the biting [\textcolor{teal}{metal}] cuffs . 
& According to the RIAA, the Beatles are the best-[\textcolor{teal}{selling}] music artists in the United States, with 178 [\textcolor{teal}{million}] certified units. They have had more number-[\textcolor{teal}{one}] albums on the [\textcolor{teal}{British}] charts and sold [\textcolor{teal}{more}] singles in the UK than any other act.\\
\midrule
\rotatebox[origin=r]{90}{RoBERTa} & Ghostbusters was [\textcolor{green}{released}] on June 8 , [\textcolor{red!75!green}{1986}] , to critical [\textcolor{green}{acclaim}] and became a cultural phenomenon . It was well [\textcolor{green}{received}] for its deft blend of comedy,  [\textcolor{green}{action}] , and horror , and Murray ' s performance was [\textcolor{red!75!green}{often}] singled out for praise .
& a filmy coating of [\textcolor{red!75!green}{dirt}] and pebbles had settled onto the block , and [\textcolor{green}{Sami}] 's hand instinctively jerked forward to swipe the [\textcolor{red}{crusty}] debris off his cheek , then pulled [\textcolor{green}{up}] short against the biting [\textcolor{red!75!green}{leather}] cuffs . 
& According to the RIAA, the Beatles are the best-[\textcolor{green}{selling}] music artists in the United States, with 178 [\textcolor{green}{million}] certified units. They have had more number-[\textcolor{green}{one}] albums on the [\textcolor{red!75!green}{US}] charts and sold [\textcolor{green}{more}] singles in the UK than any other act. \\
\midrule
\rotatebox[origin=r]{90}{Random} & Ghostbusters was [\textcolor{green}{released}] on June 8 , [\textcolor{red!75!green}{1986}] , to critical [\textcolor{green}{acclaim}] and became a cultural phenomenon . It was well [\textcolor{green}{received}] for its deft blend of comedy,  [\textcolor{green}{action}] , and horror , and Murray ' s performance was [\textcolor{red!75!green}{particularly}] singled out for praise. 
& a filmy coating of [\textcolor{red!75!green}{dirt}] and pebbles had settled onto the block , and [\textcolor{red!75!green}{Tsui}] 's hand instinctively jerked forward to swipe the [\textcolor{red}{crusty}] debris off his cheek , then pulled [\textcolor{green}{up}] short against the biting [\textcolor{red!75!green}{leather}] cuffs . 
& According to the RIAA, the Beatles are the best-[\textcolor{green}{selling}] music artists in the United States, with 178 [\textcolor{green}{million}] certified units. They have had more number-[\textcolor{green}{one}] albums on the [\textcolor{red!75!green}{US}] charts and sold [\textcolor{green}{more}] singles in the UK than any other act.
\\
\midrule
\rotatebox[origin=r]{90}{Outliers} &  \textcolor{red}{\{} \textcolor{red}{lock} was [\textcolor{red}{never}] on June 8 , [\textcolor{red}{</s>}] , to \textcolor{red}{rely} [\textcolor{red}{,}] and . It was well [\textcolor{red!75!green}{known}] for its \textcolor{red}{acker} of comedy , [\textcolor{red}{dinner}], and horror , and Murray ' s was [\textcolor{red}{ever}] \textcolor{red}{,} \textcolor{red}{</s>} \textcolor{red}{</s>} \textcolor{red}{)} 
& a \textcolor{red}{Fre ) covering} of [\textcolor{red}{humor}] and \textcolor{red}{celebcele had </s> </s> </s> </s>} , and [\textcolor{red}{</s>i}] 's \textcolor{red}{</s> </s> </s> </s> </s> </s> (@} the [\textcolor{red}{brainy}] \textcolor{red}{during (@ end)} , Then pulled [\textcolor{red}{*}] \textcolor{red}{isk ss the wearing} [\textcolor{red}{of}] cuffs </s> 
& \textcolor{red}{2017 </s>} the RIAA, the Beatles are the  [\textcolor{red}{l}] \textcolor{red}{music files} in the United States, with 178 [\textcolor{red}{Canadian}] Certified \textcolor{red}{ols </s>} They have had \textcolor{red}{é yl}-[\textcolor{red!75!green}{million}] \textcolor{red}{Deaths on} the [\textcolor{red}{Chart}] charts and \textcolor{red}{Died} [\textcolor{red}{are}] \textcolor{red}{Hearts in</s>} UK . \textcolor{red}{</s></s></s></s>} \\
\bottomrule
\end{tabular}
\caption{RoBERTa's masked language model predictions for randomly sampled input sequences. Input masked tokens (blue) are given in brackets. Correctly predicted tokens are shown in green, incorrect but plausible predictions are shown in brown. 48 weights have been modified in total for the \textit{Random} and \textit{Outliers} setups.}
\label{tab:roberta_outputs_all}
\end{table*}


\begin{table*}[]
    \footnotesize
    \centering
    \begin{tabular}{p{0.2cm}p{4.5cm}p{4.5cm}p{4.5cm}}
        \toprule
        \rotatebox[origin=r]{90}{Input} & he didnt [\textcolor{teal}{really}] have a plan and he wasnt sure he [\textcolor{teal}{could}] go through [\textcolor{teal}{with}] anything , but the [\textcolor{teal}{feeling}] of doing something was lifting his [\textcolor{teal}{spirits}] . & 
        ice is water frozen into a [\textcolor{teal}{solid}] state . [\textcolor{teal}{depending}] on the presence of impurities such as particles of soil or [\textcolor{teal}{bubbles}] of air , it can appear [\textcolor{teal}{transparent}] or a more or less [\textcolor{teal}{opaque}] bluish - white color .
        & 
        but the [\textcolor{teal}{sound}] of the river babbling by the yard and the ducks splashing on the [\textcolor{teal}{pond}] seemed to be [\textcolor{teal}{working}] a cure for her [\textcolor{teal}{melancholy}] .
        \\
        \midrule
        \rotatebox[origin=r]{90}{BERT} & he didnt [\textcolor{red!75!green}{even}] have a plan and he wasnt sure he [\textcolor{green}{could}] go through [\textcolor{green}{with}] anything , but the [\textcolor{red!75!green}{thought}] of doing something was lifting his [\textcolor{green}{spirits}] . & 
        ice is water frozen into a [\textcolor{red!75!green}{frozen}] state . [\textcolor{green}{depending}] on the presence of impurities such as particles of soil or [\textcolor{red!75!green}{particles}] of air , it can appear [\textcolor{green}{white}] or a more or less [\textcolor{red!75!green}{uniform}] bluish - white color . & 
        but the [\textcolor{green}{sound}] of the river babbling by the yard and the ducks splashing on the [\textcolor{red!75!green}{water}] seemed to be [\textcolor{red!75!green}{providing}] a cure for her [\textcolor{red!75!green}{fears}] .
        \\
        \midrule
        \rotatebox[origin=r]{90}{Random} & he didnt [\textcolor{red!75!green}{even}] have a plan and he wasnt sure he [\textcolor{green}{could}] go through [\textcolor{green}{with}] anything , but the [\textcolor{red!75!green}{thought}] of doing something was lifting his [\textcolor{green}{spirits}] . & 
        ice is water frozen into a [\textcolor{red!75!green}{liquid}] state . [\textcolor{green}{depending}] on the presence of impurities such as particles of soil or [\textcolor{red!75!green}{particles}] of air , it can appear [\textcolor{green}{white}] or a more or less [\textcolor{red!75!green}{uniform}] bluish - white color .
        & 
        but the [\textcolor{green}{sounds}] of the river babbling by the yard and the ducks splashing on the [\textcolor{red!75!green}{water}] seemed to be [\textcolor{red!75!green}{just}] a cure for her [\textcolor{red!75!green}{fears}] .
        \\
        \midrule
        \rotatebox[origin=r]{90}{Outliers} & he \textcolor{red}{didny} [\textcolor{red}{wee}] have a plan and he wasnt sure he [\textcolor{red!75!green}{would}] go through [\textcolor{red}{it}] anything , but the [\textcolor{red}{actual}] of doing something was lifting his [\textcolor{red}{shoulders}] . & 
        \textcolor{red}{that} is water turned into a [\textcolor{red}{yu}] state . [\textcolor{red!75!green}{based}] on the presence of impurities such as particles of soil or [\textcolor{red}{breath}] of air , it can appear [\textcolor{red}{white}] or a more or \textcolor{red}{more} [\textcolor{red}{commoning}] \textcolor{red}{ing} - white color .
        & but the [\textcolor{green}{sound}] of the \textcolor{red}{child} babble by the yard and the ducks splashing on the [\textcolor{red}{windows}] \textcolor{red}{all} to be [\textcolor{red}{in}] a \textcolor{red}{replacement} for her [\textcolor{red}{ness}] .
        \\
        \bottomrule
    \end{tabular}
    \caption{BERT's masked language model predictions for randomly sampled input sequences. Input masked tokens (blue) are given in brackets. Correctly predicted tokens are shown in green, incorrect but plausible predictions are shown in brown. 48 weights have been modified in total for the \textit{Random} and \textit{Outliers} setups.}
    \label{tab:bert_outputs_all}
\end{table*}

\end{document}